\documentclass[letterpaper]{article} 
\usepackage{aaai2026}  
\usepackage{times}  
\usepackage{helvet}  
\usepackage{courier}  
\usepackage[hyphens]{url}  
\usepackage{graphicx} 
\usepackage{natbib}  
\usepackage{caption} 
\usepackage{subcaption} 
\usepackage{algorithm}
\usepackage{algorithmic}

\usepackage{newfloat}
\usepackage{listings}
\DeclareCaptionStyle{ruled}{labelfont=normalfont,labelsep=colon,strut=off} 
\floatstyle{ruled}
\newfloat{listing}{tb}{lst}{}
\floatname{listing}{Listing}
\usepackage{amsmath,amsfonts,bm}

\def\gA{{\mathcal{A}}}

\def\gL{{\mathcal{L}}}

\def\gR{{\mathcal{R}}}
\def\gS{{\mathcal{S}}}

\def\gX{{\mathcal{X}}}

\newcommand{\E}{\mathbb{E}}

\newcommand{\BlackBox}{\rule{1.5ex}{1.5ex}}  
\ifdefined\proof
    \renewenvironment{proof}{\par\noindent{\bf Proof\ }}{\hfill\BlackBox\\[2mm]}
\else
    \newenvironment{proof}{\par\noindent{\bf Proof\ }}{\hfill\BlackBox\\[2mm]}
\fi
\usepackage{tikz}
\usetikzlibrary{positioning, arrows.meta, shadows, decorations.pathmorphing}
\usepackage[table]{xcolor}
\usepackage{multirow}
\usepackage{siunitx}
\usepackage{latexsym}
\usepackage{amssymb}
\usepackage{amsmath}
\usepackage{booktabs}
\usepackage{colortbl}
\usepackage{enumitem}
\usepackage{array}

\usepackage{mathtools} 
\usepackage{newunicodechar}
\newunicodechar{−}{-}

\title{Loss-Guided Auxiliary Agents for Overcoming Mode Collapse in GFlowNets}
\author{
    Idriss Malek\textsuperscript{\rm 1},
    Aya Laajil\textsuperscript{\rm 1},
    Abhijith Sharma\textsuperscript{\rm 1},
    Eric Moulines\textsuperscript{\rm 1},
    Salem Lahlou\textsuperscript{\rm 1}
}
\affiliations{
    \textsuperscript{\rm 1}Mohamed Bin Zayed University of Artificial Intelligence, UAE%
}

\usepackage{bibentry}

\begin{document}

\maketitle

\begin{abstract}
Although Generative Flow Networks (GFlowNets) are designed to capture multiple modes of a reward function, they often suffer from mode collapse in practice, getting trapped in early-discovered modes and requiring prolonged training to find diverse solutions. Existing exploration techniques often rely on heuristic novelty signals. We propose Loss-Guided GFlowNets (LGGFN), a novel approach where an auxiliary GFlowNet's exploration is \textbf{directly driven by the main GFlowNet's training loss}. By prioritizing trajectories where the main model exhibits \textbf{high loss}, LGGFN focuses sampling on poorly understood regions of the state space. This targeted exploration significantly accelerates the discovery of diverse, high-reward samples. Empirically, across \textbf{diverse benchmarks} including grid environments, structured sequence generation, Bayesian structure learning, and biological sequence design, LGGFN consistently \textbf{outperforms} baselines in exploration efficiency and sample diversity. For instance, on a challenging sequence generation task, it discovered over 40 times more unique valid modes while simultaneously reducing the exploration error metric by approximately 99\%.
\end{abstract}


\section{Introduction}
Tasks such as material design and drug discovery are fundamental to advancing essential scientific fields such as medicine \cite{Hughes2011DrugDiscovery}, energy, engineering and technology \cite{Curtarolo2013HighThroughput}. However, these tasks involve exploring vast combinatorial design spaces, ranging from $10^{60}$ possible drug-like molecules to potentially more for functional materials, making exhaustive search infeasible. Traditionally, drug and material discovery relied heavily on time-consuming and costly experimental screening and expert-guided design.  \citep{Wani1971Taxol, Klayman1985Artemisinin, Cobb2010StainlessSteel}. 

To overcome these limitations, machine learning has emerged as a key tool for accelerating discovery by providing proxy models \cite{Schutt2018Schnet, Chen2019GraphNetworks,Jumper2021AlphaFold} or methods to guide experiments \cite{Hase2018Phoenics, Hase2021Gryffin}. Moreover, it enables efficient exploration of large combinatorial spaces efficiently by leveraging their underlying structure. Reinforcement Learning (RL)~\citep{sutton1998reinforcement} is the most popular learning framework for this purpose, as it enables agents to iteratively interact with an environment to discover the single candidate that maximizes a reward, making it well-suited for tasks where the goal is to identify one optimal solution. However, in a real-world setting,  high-scoring candidates may have undesirable side effects or be synthetically inaccessible. Therefore, generating a \textbf{diverse set} of high-quality candidates is often more practical than identifying a single optimal solution, increasing the chances that at least one will be viable in practice.

Generative Flow Networks (GFlowNets) \citep{bengio2021flow} aim at learning a forward policy \( P_F(a \mid s) \), encoding choices of actions $a$ at partial states $s$, that sequentially constructs an object \( x \) with probability proportional to a given reward \( R(x) > 0 \). Unlike RL, which prioritizes finding the single best candidate, GFlowNets focus on generating a diverse set of high-reward samples. GFlowNets have found great success in many areas of applications, such as causal discovery \citep{manta2023gflownets, deleu2022bayesian}, material discovery \citep{D4DD00020J, nguyen2023hierarchical}, drug discovery \citep{shen2024tacogfn, pandey2024gflownet}, biological sequence design \citep{jain2022biological} and editing \citep{ghari2024gflownet}, large language model improvement \citep{Takase2024GFlowNet, ho2024proof, hu2024amortizing,younsi2025accurate}, diffusion models improvement \cite{DBLP:journals/corr/abs-2406-00633}, scheduling \cite{zhang2023robust} and combinatorial optimization problems \cite{zhang2023let} . In addition, alternative theories that extend GFlowNets to environments that are continuous \citep{lahlou2023theory}, stochastic \citep{pan2023stochastic}, and adversarial \citep{jiralerspong2024expected}, which further expand the scope of possible applications such as an alternative to denoising diffusion modes \cite{sendera2024improved}.

Even if GFlowNets and Reinforcement Learning have two different objectives, they share their need for exploration to achieve their goal. In sparse reward settings, an RL agent that doesn't explore will not be able to improve its ``understanding'' of the environment and will converge to a suboptimal policy. In the same manner, even though they were specifically designed to capture multiple modes of a reward function, GFlowNets training strategies that lack exploration might lead the agent to get stuck in a region and make the underlying distribution incomplete. This behavior is more visible in large sparse-reward environments \citep{malkin2023gflownets} where the agent might need to go through expansive regions with no or minimal reward.

Previous studies have aimed to enhance GFlowNet training by investigating credit assignment \citep{malkin2022trajectory, madan2023learning, pan2023better}. However, akin to RL, the effectiveness of GFlowNets relies on the trajectories used to explore high-reward regions. Due to this similarity, prior work on GFlowNets \citep{pan2023generative, rector2023thompson, lau2023dgfn, kim2023local, madantowards} has drawn insights from the RL literature \citep{chapelle2011empirical, burda2018exploration, van2016deep} to develop new exploration strategies. In particular, \citet{JMLR:v24:22-0364} briefly cite the idea of using a second GFlowNet trained mostly to match a different reward function that is high when the losses
observed by the main GFlowNet are large. Building on this idea, \citet{madantowards} suggested to implement this idea by using novelty-based intrinsic rewards, whereas \citet{kim2025adaptive} directly uses the loss value of a trajectory.

In this work, we introduce a new general training strategy for GFlowNets that can be used with any training objective. Instead of using an artificial novelty-based intrinsic reward to judge how informative a trajectory is, we directly use the loss value. To avoid ruining the objective probability distribution, we use an auxiliary agent that will sample never-seen trajectories. The Background section introduces GFlowNets, while the related works section in the appendix reviews existing exploration strategies employed in their training. Our methodology section presents the motivation for our approach, provides detailed descriptions of the proposed method, and examines the benefits of utilizing loss signals compared to conventional novelty-based intrinsic rewards. The Experiments section then shows that our method matches or exceeds the previous state-of-the-art across diverse experimental settings, including \textbf{hypergrid environments} of varying sizes and sparsity levels, \textbf{structured bit sequence generation}, \textbf{Bayesian network structure learning}, and \textbf{a novel multi-objective mRNA sequence design environment}, achieving improvements ranging from 10\% in \textbf{Bayesian structure learning} to 99\% reduction in exploration error in \textbf{sequence generation} where \textbf{prior methods failed entirely}.

The method shows \textbf{particular strength in sparse reward settings} where traditional on-policy training completely fails, as demonstrated by our motivating example where discovering a distant mode requires navigating through exponentially many low-reward states.

\section{Background}
\label{sec:Background}
\textit{Readers unfamiliar with GFlowNets are encouraged to refer to the first section of the Appendix, which provides additional context and introduces some key ideas to understand this framework.
}\\
We consider a directed acyclic graph $(\gS,\gA)$, where nodes $\gS$ represent states and edges $\gA$ represent actions. If $(s,s')\in \gA$, we say that $s$ is a parent of $s'$ and that $s'$ is a child of $s$. There is a unique state $s_0$ with no parents that we name source state and a unique state $s_f$ with no children that we name sink state. We refer to the parents of the sink state as terminating states $\gX = \{ s \in \gS \mid (s,s_f) \in \gA)\}$. A complete trajectory $\tau = (s_0, s_1,...,s_n,s_{n+1} = s_f)$ is a sequence of states that start with the source state and ends with the sink state.

We also consider a reward function $R : \gX \rightarrow \mathbf{R}$ such as $\forall x \in \gX, R(x) > 0$. By convention, we can also consider that the reward for non terminating states is zero $\forall s \in \gS \setminus \gX, R(s) = 0$. The goal of the GFlowNet is to learn to sample terminating states proportionally to the reward function $P_T(x) \propto R(x)$. To do this, GFlowNets start from the initial state (for example, empty molecule in drug discovery setting, initial coordinates in hyper-grid environment) and iteratively sample actions to move to the following states. 

Formally, we consider a non-negative flow function $F: \gA \rightarrow \mathbf{R}$. The goal of the GFlowNet framework is to learn such a flow function that satisfies the following flow matching (FM) and reward matching constraints, for all $s' \neq s_0,s_f$ and for all $x \in \gX$, respectively:
\begin{align}
    &\sum_{(s,s')\in\gA} F(s \rightarrow s') = \sum_{(s',s'')\in\gA} F(s' \rightarrow s''),\\
    &F(x \rightarrow s_f) = R(x).
\end{align}
We extend the definition of the flow function to states:
\begin{align}
    \forall s \in \gS,\quad F(s) = \sum \limits_{(s,s')\in\gA} F(s \rightarrow s').
\end{align}
We also define the forward and backward policies as follow:
\begin{align}
    P_F(s'\mid s) = \frac{F(s \rightarrow s')}{F(s)}, \ \
    P_B(s\mid s') = \frac{F(s \rightarrow s')}{F(s')}.
\end{align}

In this paper, we use the trajectory balance (TB) objective to train GFlowNets. Introduced in \citet{malkin2022trajectory}, the trajectory-decomposable loss only parametrizes the forward $P^\theta_F$  and backward $P^\theta_B$ policies, and adds a learnable scalar $Z_\theta$ that represents the unknown partition function. The loss for each trajectory $\tau$:
\begin{align}
    \scalebox{0.97}{$\gL_{TB}(\tau; \theta) =
\left( \log \left( \frac{Z_\theta \prod_{t=1}^{n+1} P^\theta_F(s_t \mid s_{t-1})}
{R(s_n) \prod_{t=1}^{n} P^\theta_B(s_{t-1} \mid s_t)} \right) \right)^2,$}
\end{align}
is minimized by stochastic gradient descent, using trajectories sampled from a given behavior policy. Common choices of the behavior policy include the learned policy $P^\theta_F$ and artificially modified versions of it~\citep{bengio2023gflownet}. In this work, we tackle the problem of behavior policy design, which we discuss next.

\section{Methodology}

In this section, we describe our methodology for improving the training efficiency and exploration capabilities of GFlowNets. We begin by illustrating a minimal environment that highlights the limitations of on-policy GFlowNet training in discovering high-reward modes. Building on this insight, \textbf{we introduce a novel training framework that leverages an auxiliary GFlowNet guided by the main GFlowNet's loss to promote exploration in underrepresented regions of the state space}. Finally, we discuss the practical benefits of this loss-guided auxiliary mechanism, particularly in settings where neural network generalization and efficient credit assignment are critical.

\label{sec:Methods}
\subsection{Motivation}

\begin{figure}[H]
\centering
\begin{tikzpicture}[>=latex, ->, line width=0.7pt]

  \tikzstyle{extremal} = [circle, draw=black, fill=teal!50, minimum size=1.2cm, font=\boldmath]
  \tikzstyle{intermediate} = [circle, draw=black, fill=teal!10, minimum size=1.2cm, font=\boldmath]

  \node[extremal] (s0) at (0, 0) {$s_0$};
  \node[intermediate] (s1) at ([xshift=1.8cm]s0) {$s_1$};
  \node (sdots) at ([xshift=1.8cm]s1) {\Large$\cdots$};
  \node[intermediate] (sNm1) at ([xshift=1.8cm]sdots) {$s_{N-1}$}; 
  \node[extremal] (sN) at ([xshift=1.8cm]sNm1) {$s_N$};

  \draw (s0) -- (s1);
  \draw (s1) -- (sdots);
  \draw (sdots) -- (sNm1);
  \draw (sNm1) -- (sN);

\end{tikzpicture}
\caption{Directed chain of \( N+1 \) states. Extremal nodes \(s_0\) and \(s_N\) have high reward, while all intermediate states have low reward.}
\label{fig:state_chain}
\end{figure}

\label{sec:motiv}

To understand why GFlowNets can fail to learn the desired distribution, we consider a simple illustrative environment, depicted in Figure \ref{fig:state_chain}. The state space is defined as $\gS = \{s_0, s_1, \dots, s_N\} \cup {s_f}$, and the action space is $\gA = \{\texttt{advance}, \texttt{exit}\}$. For every $i \leq N$, the action \texttt{exit} leads from $s_i$ to the terminal state $s_f$, and for every $i < N$, the action \texttt{advance} leads from $s_i$ to $s_{i+1}$. The reward function assigns high values to the extremal states, with $R(s_0) = R(s_N) \gg R(s_i)$ for all $i \in {1, \dots, N-1}$. We use a tabular GFlowNet where each state $s$ is mapped to a parameter $\theta_s$ such that $P_F(\texttt{exit}|s) = \sigma(\theta_s)$ (where $\sigma$ is the sigmoid function) and a parameter $Z$ that represents the total reward learned. 

\begin{figure}[H]
    \centering
    \includegraphics[width=\columnwidth]{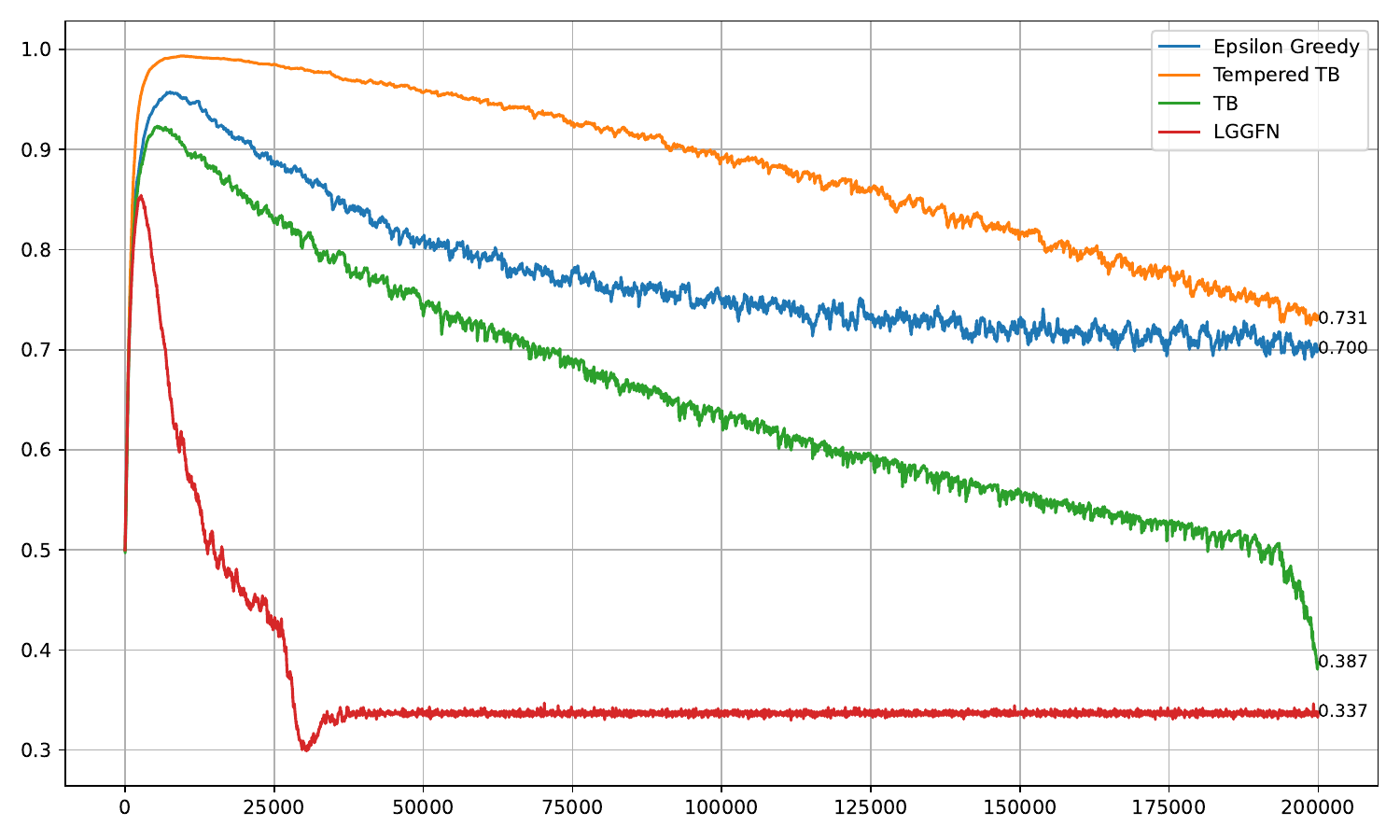}
    \caption{Plot of the evolution of $P_F(\texttt{exit} \mid s_0)$ for different algorithms as a function of number of sampled trajectories. The chain has 100 states and uses the reward setup $R(s_0) = R(s_N) = 101$ and $R(s_i) = 1$. At convergence,  $P_F(\texttt{exit} \mid s_0) = \frac{101}{300} \approx 0.337$.}
    \label{fig:chain_plot}
\end{figure}

Despite its simplicity (linear, non-branching DAG structure with a single non-terminal action and binary reward pattern), on-policy GFlowNet training fails to converge within a reasonable number of iterations. Intuitively, for the GFlowNet to approximate the correct distribution, it must discover and propagate the high reward associated with $s_N$. However, assuming the initial policy is a uniform action policy where each action is chosen with probability 0.5, the probability of reaching $s_N$ is $0.5^N$. This probability diminishes exponentially with $N$, and as training progresses, the policy increasingly prioritizes the more easily discovered high-reward state $s_0$, further reducing the chance of ever reaching $s_N$. As a result, $s_N$ is typically discovered far later than expected (approximately 190,000 trajectories in the case of a chain of length 100, as indicated by a noticeable inflection point in the green curve in Figure~\ref{fig:chain_plot}), which significantly slows down learning despite the simplicity of the environment.

In more complex settings where rewards vary more widely, a similar phenomenon can occur: trajectories may get trapped in suboptimal regions around locally high-reward states, further impeding exploration and slowing down convergence.

\subsection{Algorithm}
In the example illustrated in Figure~\ref{fig:state_chain}, once the model has learned to terminate trajectories at $s_0$, such trajectories become dominant on policy, making the discovery of the final state slower. Figure~\ref{fig:chain_plot} shows that even after 200,000 trajectories, training using an on-policy method (TB) has yet to converge for a 100-sized chain. Moreover, standard exploration techniques such as tempering or $\varepsilon$-greedy exploration prove insufficient in overcoming this limitation.

To address this, we propose \textbf{training the main GFlowNet using trajectories sampled from an auxiliary GFlowNet}, which is optimized to sample trajectories according to a modified reward:
\[
R_{\text{aux}} = R_{\text{main}} + \lambda \mathcal{L}_{\text{main}},
\]
where $R_{\text{main}}$ is the original reward used to train the main GFlowNet, and $\mathcal{L}_{\text{main}}$ denotes the loss associated with a state or trajectory under the main GFlowNet. This auxiliary reward prioritizes trajectories that the main GFlowNet has not yet learned (i.e., those with high loss), while reducing focus on already well-learned regions (i.e., low-loss trajectories). While we mainly focus on trajectory-based objectives in this paper, the same idea holds for transition, state or sub-trajectory -based objective. As a result, this encourages broader exploration and mitigates \textbf{mode collapse}.

To leverage the auxiliary GFlowNet, we train the main agent using a mixture of on-policy trajectories and trajectories sampled by the auxiliary GFlowNet. This strategy helps the main agent retain knowledge of previously discovered modes while also exposing it to regions where its performance remains weak. To mitigate the training instability of the auxiliary gflownet that is due to a moving reward, we include $R_{\text{main}}$ in the auxiliary reward to provide a consistent structural signal, thereby counteracting the inherent fluctuations of the loss term. The coefficient $\lambda$ balances the influence of the loss and the reward, ensuring that the overall scale of $R_{\text{aux}}$ remains comparable to that of $R_{\text{main}}$ and avoids destabilizing the learning process. The training procedure is outlined in detail in Algorithm \ref{alg:auxiliary_gflownet}.
\begin{algorithm}[t]
\caption{Training with Auxiliary GFlowNet}
\label{alg:auxiliary_gflownet}
\begin{algorithmic}[1]
\STATE \textbf{Initialize:} Main GFlowNet $F_{\text{main}}$, Auxiliary GFlowNet $F_{\text{aux}}$, reward function $R$, auxiliary weight $\lambda$

\FOR{each training iteration}
    \STATE Sample trajectories $\tau_{\text{aux}} \sim F_{\text{aux}}$, with reward $R$
    \STATE Sample trajectories $\tau_{\text{main}} \sim F_{\text{main}}$
    \STATE Compute main loss $\mathcal{L}(\tau_{\text{aux}}; F_{\text{main}}, R)$
    \STATE Compute auxiliary reward: $R_{\text{aux}} = R + \lambda \mathcal{L}(\tau_{\text{aux}}; F_{\text{main}}, R)$
    \STATE Update $F_{\text{aux}}$ using $\mathcal{L}(\tau_{\text{aux}}; F_{\text{aux}}, R_\text{aux})$
    \STATE Concatenate trajectories: $\tau = \tau_{\text{main}} \cup \tau_{\text{aux}}$
    \STATE Update $F_{\text{main}}$ using $\mathcal{L}(\tau; F_{\text{main}}, R)$
\ENDFOR
\end{algorithmic}
\end{algorithm}

\subsection{Benefits of loss-guided auxiliary GFlowNet}
\label{sec:benefits}
Using the loss of the main GFlowNet as a guide for the auxiliary GFlowNet allows to leverage the generalization induced by the use of Neural Networks.
During training, neural networks learn to recognize patterns and generalize them to previously unseen states. For instance, as discussed in the BitSequence subsection in the experiments, the model may infer that when a sequence is incomplete but valid, the optimal next action leading to high-reward states is consistently appending a $0$ (an open parentheses). Due to this generalization, some trajectories or transitions, despite never being explicitly encountered, can still produce a low loss. Leveraging the loss to guide exploration takes advantage of this property. 

In contrast, relying solely on novelty-based intrinsic rewards might drive the model toward areas that are technically new but were already implicitly understood through the network's generalization. Using the loss is also cheaper computationally, since it will be computed anyway to train the main agent, whereas intrinsic rewards ask for a separate set of calculations, and can be expensive depending on which novelty-based algorithm is used.

While similar in spirit, the reward formulation proposed in \citet{kim2025adaptive} differs in two aspects. First, it is asymmetric, placing greater emphasis on trajectories where the estimated forward reward $ZP_F(\tau)$ is lower than the expected reward $R(x_\tau)P_B(\tau \mid x_\tau)$. Second, it combines the primary and auxiliary objectives multiplicatively, using the formulation $\log \gR_{\text{aux}} = \alpha \log \gR + \log \gL$. However, our experiments indicate that this added complexity does not lead to improved performance. In fact, our simplified variant outperforms it by a small margin. 

A short theoretical discussion of why loss-guided sampling prevents partial-support stationary points and how it implicitly induces a curriculum over trajectories is provided in the Appendix.

\section{Experiments}
\label{sec:Experiments}
 In the experiments section, we compare three GFlowNets training procedures with the trajectory balance objective: on-policy (that we simply refer to as TB), Siblings Augmented GFlowNet \citep[SAGFN;][]{madantowards}, and our loss-guided auxiliary GFlowNet (LGGFN). We first validate the ability of both SAGFN and LGGFN to discover all modes in the hyper-grid environemnt introduced in \citet{bengio2021flow}, then we show that LGGFN outperforms SAGFN in settings where the rewards have a strong structure: a bit-sequence environment similar to \cite{malkin2022trajectory}'s, a causal structure learning environment \citep{deleu2022bayesian}, and a structured biological sequence design setting..
\subsection{Hypergrid}
The Hypergrid environment, introduced in \cite{bengio2021flow}, is a standard setup in GFlowNet literature. In this environment, states are represented as integer vectors corresponding to points in a grid, and actions involve incrementing one of the coordinates by $1$, starting from $s_0 = \bm{0}$. The reward function typically assigns higher values to the corners of the grid (for more details on the environment, refer to the Appendix).
To showcase that getting stuck in a mode is one of the main reasons GFlowNets fail to explore the state space, we present a thorough experimentation on the Hypergrid environment.
\subsubsection{Comparison between LGGFN and adaptive teachers:}
As discussed in the Methodology section, \citet{kim2025adaptive} propose a related approach. In this section, we present experimental results (Table \ref{tab:hypergrid_adaptive}) demonstrating that our simpler method performs on par with their more complex formulation. Therefore, for the remainder of our experiments, we adopt our version of the loss-guided auxiliary GFlowNet.
\begin{table}[H]
\centering

\setlength{\tabcolsep}{2pt}
\renewcommand{\arraystretch}{1.3}

\begin{tabular}{llccc}
\toprule
\multirow{2}{*}{\textbf{Method}} & \multicolumn{3}{c}{\textbf{Grid Size}} \\
\cmidrule(lr){2-4}
& \textbf{32×32} & \textbf{64×64} & \textbf{128×128} \\
\midrule

Adaptive Teachers & $7.20 \pm 2.00$ 
        & $2.20 \pm 0.31$ 
        & $0.92 \pm 0.36$ \\
\rowcolor{teal!15}
LGGFN & $\mathbf{5.33} \pm 1.89$ 
         & $\mathbf{2.13} \pm 0.61$ 
         & $\mathbf{0.83} \pm 0.21$ \\
\bottomrule
\end{tabular}
\caption{Final $\ell_1$ distance (scaled by \( \times 10^{-5} \)) on Hypergrid benchmarks after $10^4$ sampled trajectories. LGGFN outperforms Adaptive Teachers \cite{kim2025adaptive} across grid sizes. Hyperaparameters are given in the appendix.}
\label{tab:hypergrid_adaptive}
\end{table}
Additionally, we observed that the performance and stability of \cite{kim2025adaptive} were highly dependent on the selection of hyperparameters. Specifically, deviations from the hyperparameter values specified in the original work frequently led to numerical instability and NaN values, particularly for larger grid sizes (128 x 128). In contrast, our version exhibits a significantly lower sensitivity to hyperparameter variations, as demonstrated later in the Experiments.

\subsubsection{Different sizes:}
As shown by the toy example, the size of the environment exponentially  impacts the ability of the model to escape explored modes. We validate this observation with a much more complete experimentation (Table \ref{tab:hypergrid_transposed_scaled}). In the following experiments, we use a sparse reward parameter that makes the exploration of modes far from $s_0$ harder ($R_0 = 0.0001$, see the Appendix for more details).

\begin{figure*}[ht]
    \centering
    \includegraphics[width=1.\textwidth]{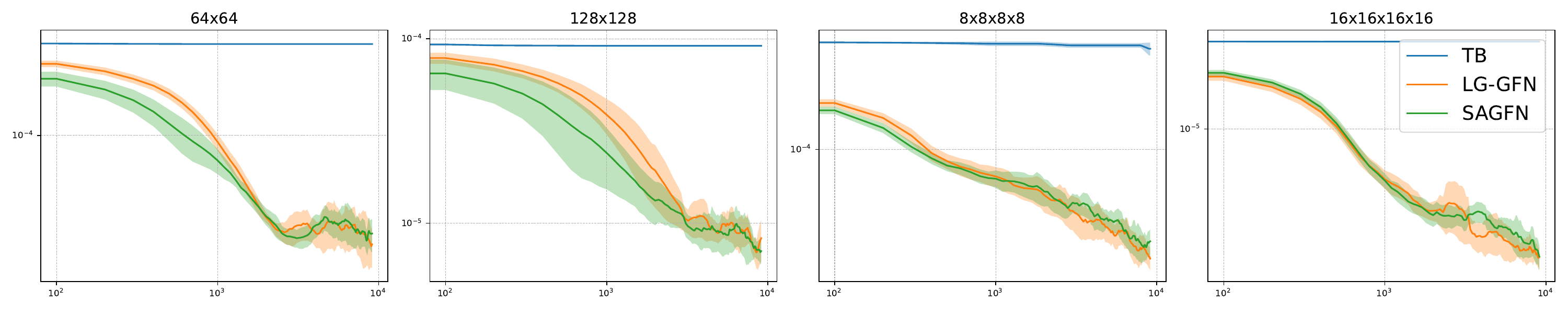}
    \caption{L1-loss during training for different sizes of hypergrid as a function of training iterations.}
    \label{fig:hypergrid_plots}
\end{figure*}

We observe that on-policy training quickly becomes ineffective as the size of the hypergrid increases, as it struggles to escape the first mode within a reasonable amount of time (see the appendix for visualizations of the learnt distributions). In contrast, both SAGFN and LGGFN perform similarly on this task: they consistently discover all the modes and achieve nearly identical L1 losses across all experiments. For these two algorithms, we can distinguish two distinct training phases in Figure \ref{fig:hypergrid_plots}. The first phase is characterized by a steep decline in the loss curve, indicating active exploration and significant influence from the auxiliary GFlowNet's loss or intrinsic reward. The second phase begins when the curve starts to plateau, with only slow, incremental improvement: at this point, the loss or intrinsic reward has diminished to the extent that switching to on-policy training would yield comparable results. It is also worth noting that LGGFN shows more consistent training behavior during the first phase across experiments with different random seeds, as evidenced by a smaller standard deviation (reflected by the narrower standard deviation band in the plot) compared to SAGFN.
\begin{figure*}[ht]
    \centering
    \includegraphics[width=1.\textwidth]{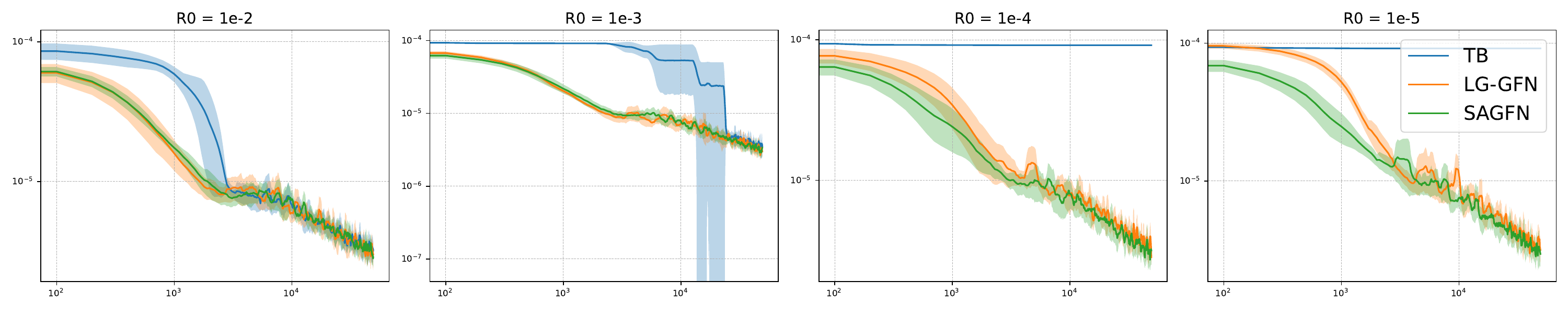}
    \caption{L1-loss during training for different values of $R_0$ and a fixed size of 128x128. as a function of training iterations.}
    \label{fig:hypergrid_plots_sparsity}
\end{figure*}

\begin{table}[H]
\centering

\setlength{\tabcolsep}{7pt}
\renewcommand{\arraystretch}{1.3}

\resizebox{0.95\linewidth}{!}{%
\begin{tabular}{lcc>{\columncolor{teal!15}}c}
\toprule
& \textbf{TB} & \textbf{SAGFN} & \textbf{LGGFN} \\
\cmidrule{2-4}
\textbf{Size} & \multicolumn{3}{c}{\textit{2-Dimensional Grid}} \\
\midrule
$32$       & $1460.00 \pm 0.016$   & $63.9 \pm 12.4$     & $\mathbf{53.3 \pm 18.9}$ \\
$64$       & $366.00 \pm 0.010$    & $24.8 \pm 5.64$     & $\mathbf{21.3 \pm 6.14}$ \\
$80$       & $234.00 \pm 0.0039$   & $17.5 \pm 4.51$     & $\mathbf{15.7 \pm 3.44}$ \\
$96$       & $163.00 \pm 0.0070$   & $11.4 \pm 3.09$     & $\mathbf{10.6 \pm 1.95}$ \\
$128$     & $91.6 \pm 0.0008$     & $\mathbf{7.07 \pm 1.00}$ & $8.30 \pm 2.10$ \\
$144$     & $72.3 \pm 0.0009$     & $2.71 \pm 0.493$    & $\mathbf{2.57 \pm 0.373}$ \\
\midrule
& \multicolumn{3}{c}{\textit{4-Dimensional Grid}} \\
\midrule
$4$     & $113.0 \pm 39.9$      & $169.0 \pm 47.7$    & $\mathbf{113.0 \pm 58.9}$ \\
$8$     & $417.0 \pm 38.2$      & $27.1 \pm 5.04$     & $\mathbf{21.2 \pm 3.19}$ \\
$16$ & $28.6 \pm 0.000099$   & $2.15 \pm 0.314$    & $\mathbf{2.13 \pm 0.309}$ \\
\bottomrule
\end{tabular}
}
\caption{Final $\ell_1$ distance (scaled by \( \times 10^{-6} \)) for various Hypergrid sizes after $10^4$ sampled trajectories. Lower is better.}
\label{tab:hypergrid_transposed_scaled}
\end{table}

\begin{table*}[ht]
\centering

\setlength{\tabcolsep}{1pt}
\renewcommand{\arraystretch}{1.3}

\resizebox{\textwidth}{!}{%
\begin{tabular}{l *{5}{>{\centering\arraybackslash}m{1.8cm} >{\centering\arraybackslash}m{1.8cm}}}
\toprule
\multirow{2}{*}{\textbf{Method}} 
& \multicolumn{2}{c}{\textbf{16}} 
& \multicolumn{2}{c}{\textbf{24}} 
& \multicolumn{2}{c}{\textbf{32}} 
& \multicolumn{2}{c}{\textbf{40}} 
& \multicolumn{2}{c}{\textbf{48}} 
\\
\cmidrule(lr){2-3} \cmidrule(lr){4-5} \cmidrule(lr){6-7} \cmidrule(lr){8-9} \cmidrule(lr){10-11}
& \textbf{Diversity} & \textbf{Exploration} 
& \textbf{Diversity} & \textbf{Exploration} 
& \textbf{Diversity} & \textbf{Exploration} 
& \textbf{Diversity} & \textbf{Exploration} 
& \textbf{Diversity} & \textbf{Exploration} 
\\
\midrule
TB 
  & 23 $\pm$ 3 & 0.92 $\pm$ 0.00
  & \textbf{15309} $\pm$ 40 & \textbf{0.01} $\pm$ 0.00
  & 23 $\pm$ 3 & 0.92 $\pm$ 0.00
  & 89 $\pm$ 7 & 0.99 $\pm$ 0.00
  & 54 $\pm$ 5 & 0.99 $\pm$ 0.00
\\
SAGFN 
  & 473 $\pm$ 640 & 0.63 $\pm$ 0.41  
  & 7387 $\pm$ 95 & 0.05 $\pm$ 0.01 
  & 4392 $\pm$ 3129 & 0.45 $\pm$ 0.39
  & 102 $\pm$ 13 & 0.98 $\pm$ 0.01 
  & 35 $\pm$ 8 & 0.99 $\pm$ 0.01
\\
\rowcolor{teal!15}
LGGFN 
  & \textbf{1425} $\pm$ 2 & \textbf{0.01} $\pm$ 0.01 
  & 7770 $\pm$ 10 & 0.01 $\pm$ 0.01
  & \textbf{7805} $\pm$ 39 & \textbf{0.02} $\pm$ 0.01 
  & \textbf{7746} $\pm$ 53 & \textbf{0.02} $\pm$ 0.01 
  & \textbf{7553} $\pm$ 65 & \textbf{0.05} $\pm$ 0.01 
\\
\bottomrule
\end{tabular}
}
\caption{Comparison of on-policy (TB), SAGFN, and LGGFN across different sequence sizes.}
\label{tab:parentheses}

\end{table*}

\begin{table*}[ht]
\centering

\setlength{\tabcolsep}{6pt}
\renewcommand{\arraystretch}{1.25}

\resizebox{\textwidth}{!}{%
\begin{tabular}{lccccccccc}
\toprule
\multirow{3}{*}{\textbf{Method}} & \multicolumn{9}{c}{\textbf{Graph Size (Nodes)}} \\
\cmidrule(lr){2-10}
& \textbf{5} & \textbf{6} & \textbf{7} & \textbf{8} & \textbf{9} & \textbf{10} & \textbf{11} & \textbf{12} & \textbf{13} \\
\cmidrule(lr){2-10}
& \multicolumn{9}{c}{\textit{ROC AUC}} \\
\midrule
TB & \textbf{0.72} $\pm$ 0.08 & 0.57 $\pm$ 0.15 & 0.58 $\pm$ 0.08 & \textbf{0.60} $\pm$ 0.06 & 0.53 $\pm$ 0.07 & 0.53 $\pm$ 0.13 & 0.56 $\pm$ 0.09 & 0.52 $\pm$ 0.08 & 0.56 $\pm$ 0.05 \\
SAGFN & 0.66 $\pm$ 0.15 & \textbf{0.68} $\pm$ 0.10 & 0.44 $\pm$ 0.07 & 0.51 $\pm$ 0.16 & \textbf{0.56} $\pm$ 0.13 & 0.59 $\pm$ 0.06 & 0.49 $\pm$ 0.03 & 0.52 $\pm$ 0.08 & 0.55 $\pm$ 0.04 \\
\rowcolor{teal!15}
LGGFN & 0.57 $\pm$ 0.18 & \textbf{0.68} $\pm$ 0.05 & \textbf{0.62} $\pm$ 0.16 & 0.52 $\pm$ 0.08 & \textbf{0.57} $\pm$ 0.05 & \textbf{0.62} $\pm$ 0.07 & \textbf{0.63} $\pm$ 0.11 & \textbf{0.59} $\pm$ 0.03 & \textbf{0.57} $\pm$ 0.07 \\
\midrule
& \multicolumn{9}{c}{\textit{$\mathbb{E}$-SHD}} \\
\midrule
TB & 7.33 $\pm$ 1.08 & 12.28 $\pm$ 1.06 & 16.69 $\pm$ 0.12 & \textbf{20.25} $\pm$ 2.62 & 29.52 $\pm$ 2.11 & 37.31 $\pm$ 3.97 & 42.60 $\pm$ 3.38 & 54.09 $\pm$ 0.41 & 64.20 $\pm$ 3.05 \\
SAGFN & 7.62 $\pm$ 0.23 & \textbf{11.13} $\pm$ 0.85 & 18.03 $\pm$ 0.49 & 23.19 $\pm$ 2.20 & \textbf{29.34} $\pm$ 2.60 & \textbf{36.53} $\pm$ 1.12 & 45.58 $\pm$ 0.76 & 52.88 $\pm$ 4.78 & 63.31 $\pm$ 3.77 \\
\rowcolor{teal!15}
LGGFN & \textbf{7.12} $\pm$ 1.38 & 11.83 $\pm$ 0.37 & \textbf{15.81} $\pm$ 2.47 & 23.78 $\pm$ 1.36 & \textbf{29.11} $\pm$ 1.17 & \textbf{36.68} $\pm$ 0.63 & \textbf{41.33} $\pm$ 3.50 & \textbf{51.77} $\pm$ 3.50 & \textbf{63.00} $\pm$ 4.50 \\
\bottomrule
\end{tabular}
}
\caption{Comparison of on-policy (TB), SAGFN, and LGGFN on ROC AUC and SHD metrics across different graph sizes.}
\label{tab:bayesian}

\end{table*}

\subsubsection{Different sparsity:}
In the toy example, the main reason the GFlowNet couldn't access the last high reward state is the low reward of the intermediate states, that led to a low move-on probability in the initial state. To validate this hypothesis, we evaluate a range of reward configurations with varying levels of sparsity by systematically modifying the value of $R_0$ (Figure \ref{fig:hypergrid_plots_sparsity}).

We observe that, even in relatively easy settings, on-policy training results in delayed convergence. Furthermore, once all modes have been discovered, the behavior of all three algorithms converges, suggesting that the auxiliary agent can be discarded at this stage. This allows continued on-policy training without performance degradation, while reducing computational overhead.

\subsubsection{Influence of the coefficient $\lambda$:}

\label{sec:hyperparameter_grid_comparison}
\begin{figure}[H]
    \centering
    \includegraphics[width=\linewidth]{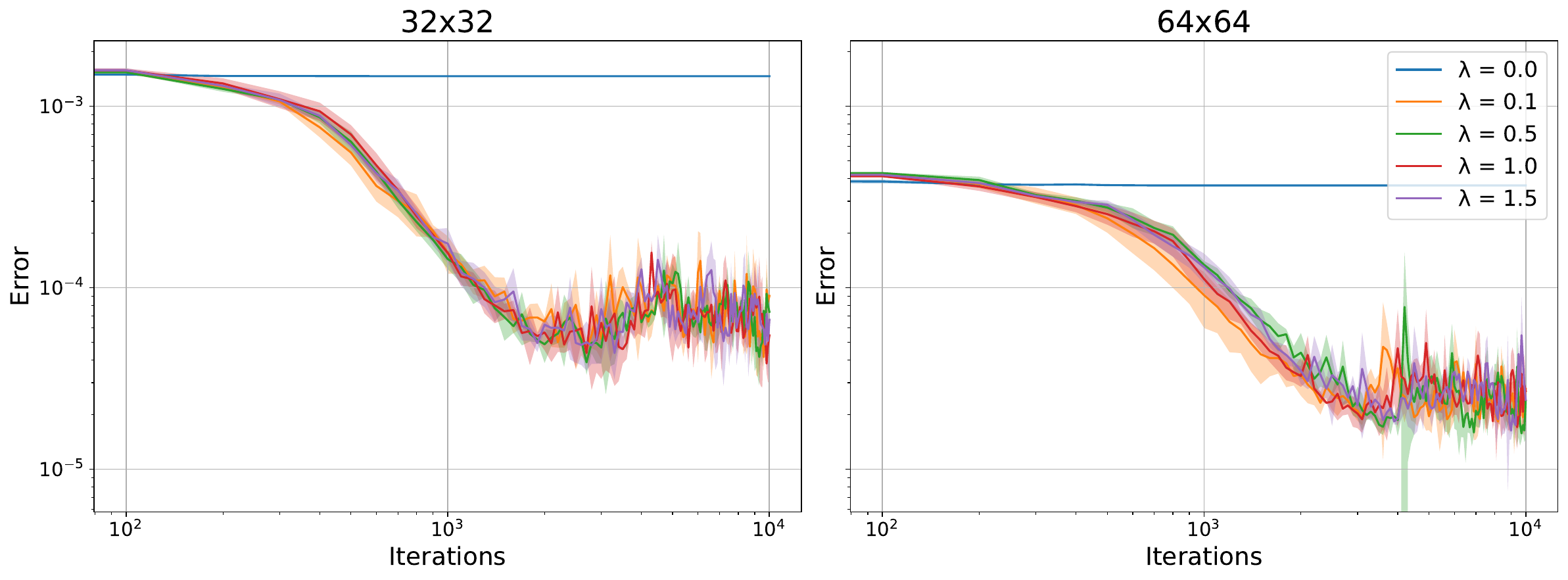}
    \caption{L1-loss evolution over training iterations for different $\lambda$ values and grid sizes.}
    \label{fig:hyperparameter_grid_comparison}
\end{figure}

The auxiliary reward incorporates a coefficient $\lambda$ to balance the original reward $R$ and the additional loss term $\mathcal{L}$. However, experimental results indicate that when $\mathcal{L}$ is initially scaled to be in the same range as $R$, variations in $\lambda$ within a reasonable range (i.e., not approaching zero) have negligible effect on performance, as demonstrated in Figure~\ref{fig:hyperparameter_grid_comparison}.

\subsection{Valid bit sequences}
\label{sec:bitsequence}
The \textsc{BitSequences} environment was introduced as a testbed for studying structured sequence generation by \citet{malkin2022trajectory}. Each state in the environment corresponds to a binary sequence, and the agent's actions consist of appending either individual bits or blocks of bits to the current sequence. Episodes start with an empty sequence $s_0 = \epsilon $ and terminate when the sequence reaches a predefined length. In the original formulation, the reward function quantifies how close a given sequence is to the set of modes $R(s) = e^{-d(s, \text{modes})}$. This set of modes is constructed by first arbitrarily selecting a set of words $H$ and then generating different complete sequences by randomly combining these words. \citet{malkin2022trajectory} argues that this construction inherently imposes a structural property on the reward function. However, this claim is not entirely accurate, as the candidate sequences that were not ultimately selected as modes still share the same underlying structure but may receive a significantly lower reward. For example, if we work on sequences of size 6 and the set of arbitrary words is $H = \{ 0 1, 1 0\} $ and the set of modes is $\{010101, 101010, 100110\}$. In this setting, the sequence $011001$ shares the same "structure" of the modes as it is built from the same process but has a much lower reward $e^{-2}$ (the reward of the modes being 1). \\
We introduce a new reward structure for this environment (the motivation behind this modification is given in the Appendix). The set of valid sequences is defined recursively as the smallest set that contains the empty sequence, is closed under concatenation, and is stable under the simultaneous operation of prepending a $0$ and appending a $1$. Interpreting $0$ as an opening parenthesis \texttt{(} and $1$ as a closing parenthesis \texttt{)}, a valid sequence corresponds to a balanced parentheses sequence. When considering sequences of maximum length $2N$, we define a sequence as \textit{complete} if it attains this maximum length. The objective of the task is to train a GFlowNet to discover the full set of complete valid sequences.

To evaluate the performance of different algorithms in this environment, we sample 16,000 sequences and assess them using two key metrics:

\begin{itemize}
    \item \textbf{Exploration~$\downarrow$:} We estimate the probability mass assigned to the set of complete valid sequences using Monte Carlo sampling and report its difference to the true probability mass.
    \item \textbf{Diversity~$\uparrow$:} We measure the fraction of distinct complete valid sequences sampled. For sequences of length $2N$, the number of possible valid bit sequences is given by the $N$-th catalan number:
    \begin{equation}
        C_N = \frac{1}{N+1} \binom{2N}{N} = \frac{(2N)!}{(N+1)!N!}.
    \end{equation}
\end{itemize}

The results are presented in Table~\ref{tab:parentheses}. Overall, LGGFN significantly outperforms the other algorithms, particularly on longer sequences. Its advantage over SAGFN is likely attributable to its ability to leverage generalization, as hypothesized earlier. Architectural details and hyperaparameters are provided in the Appendix.

\subsection{Bayesian structure learning}
Bayesian networks~\cite{pearl2014probabilistic} are probabilistic graphical models that represent a set of variables and their conditional dependencies using a directed acyclic graph (DAG). When this graph is not known \textit{a priori} (e.g., from expert knowledge), it can be inferred from a dataset of observations $\mathcal{D}$.

Traditional algorithms for this task typically return a single DAG~\citep{chickering2002optimal}, which is suboptimal in practice. One reason for this limitation is that such methods fail to account for the inherent uncertainty arising from Markov equivalence (explained in the Appendix). 

Graphs within the same equivalence class are therefore indistinguishable solely on observational data. Another important consideration is that, when the dataset $\mathcal{D}$ is limited, it may not provide sufficient evidence to clearly favor a particular equivalence class. In such cases, predicting a single graph (or a single equivalence class) may lead to poor calibration. Hence, it is more appropriate to estimate a posterior distribution over DAGs, $P(G\mid\mathcal{D})$, which better captures the range of plausible graph structures supported by the data.

The use of GFlowNets for Bayesian structure learning was previously explored by \citet{deleu2022bayesian}, yielding strong empirical results. Building on this work, we evaluate the effectiveness of various algorithms in the context of Bayesian structure learning. Details of the environment, architectures, and hyperparameters, are in the appendix.

We evaluate the models using two metrics: \begin{itemize} \item \textbf{Expected Structural Hamming Distance} ($\E$-SHD)~$\downarrow$: the number of missing or extra edges, and the number of reversals required to transform one graph into another. \item \textbf{ROC-AUC}~$\uparrow$: measures how well the predicted edge probabilities distinguish real edges from non-existent edges. \end{itemize}

Experimental results are presented in Table~\ref{tab:bayesian}. For graphs with a small number of nodes (5 to 8), all three algorithms exhibit competitive performance, with no single method consistently outperforming the others. However, LGGFN demonstrates a clear advantage as the graph size increases (9 nodes and above). Given that the size of the state space grows rapidly with the number of nodes (see the Appendix), this performance gap is likely due to LGGFN's ability to leverage generalization effectively. By learning to identify and avoid unpromising regions of the search space, LGGFN reduces the need for exhaustive exploration, directly leading to better results.

\subsection{Structured Biological Sequence Generation}
Designing messenger RNA (mRNA) sequences that efficiently and stably express target proteins is a central challenge in synthetic biology and medicine \cite{zhang2023algorithm, fallahpour2025codontransformer}. Due to the redundancy of the genetic code, each protein can be encoded by an exponentially large number of synonymous mRNA sequences, which makes this task well suited to GFlowNet-based generative modeling~\cite{jain2022biological,cretu2024synflownet}. However, not all synonymous sequences are equally effective biologically. We define a multi-objective reward function \cite{jain2023multi} that captures desirable biological properties. We create a generative framework that learns to generate synonymous codon sequences while adhering to user-customized biological constraints. The objective involves the Codon Adaptation Index (CAI), the GC content, and the Minimum Free energy (MFE), similar to \citet{laajil2025curriculum}, all explained in the Appendix.

\subsubsection{Results:}  
We evaluated three GFlowNet variants, TB, LGGFN, and SAGFN, for designing mRNA sequences encoding the protein, \textit{Acyl-CoA thioesterase 13} (ACOT13), a 151-amino acid enzyme involved in lipid metabolism. LGGFN demonstrated superior multi-objective mRNA design by generating sequences with higher GC content (up to 55.11\% vs.\ 45.33\% natural), more stable RNA secondary structures (lowest MFE of $-91.30$ kcal/mol vs.\ $-68.20$ kcal/mol natural), and improved codon adaptation index (CAI up to 0.62 vs.\ 0.54 natural). The generated sequences also exhibited substantial diversity, as measured by Levenshtein distance between the generated sequences averaging around 50–60 from each other.

\begin{figure}[h!]
    \centering
    \begin{subfigure}[b]{0.5\linewidth}
        \centering
        \includegraphics[width=\linewidth]{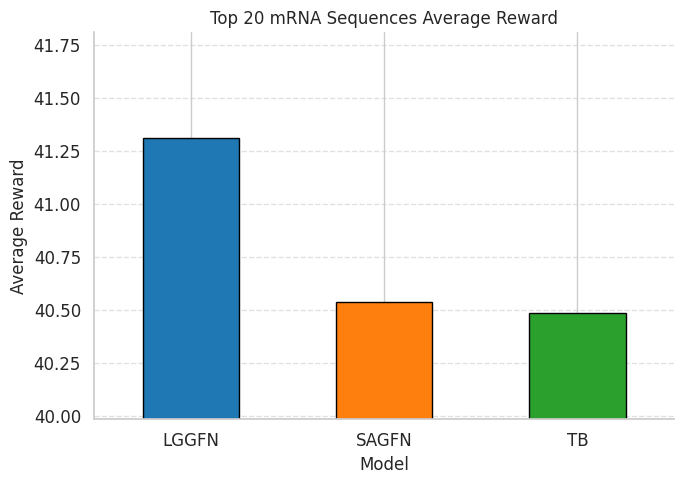}
        \label{fig:seq_metrics}
    \end{subfigure}
    \hfill
    \begin{subfigure}[b]{0.49\linewidth}
        \centering
        \includegraphics[width=\linewidth]{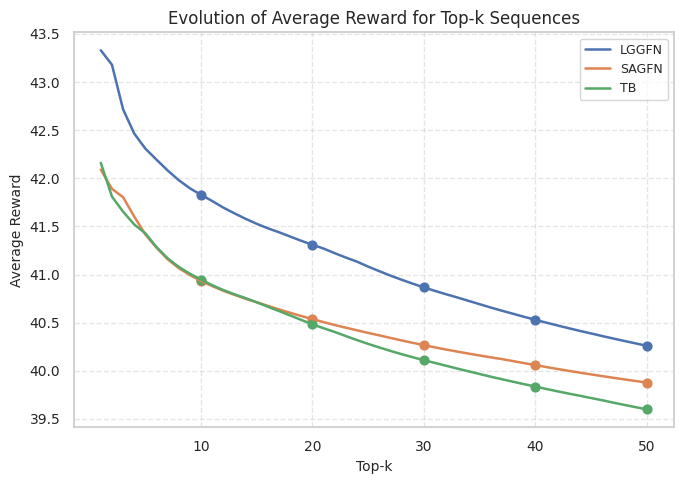}
        \label{fig:evolution_avg_reward}
    \end{subfigure}
    \caption{We find that LGGFN [\textbf{Left}] generates the highest average reward among the top 20 sequences compared to SAGFN and TB, and [\textbf{Right}] consistently outperforms others across the evolution of the average reward as a function of the top-k sequences per model for k = [10,20,30,40,50].}
    \label{fig:combined_metrics}
\end{figure}

\section{Conclusion}
\label{sec:Conclusion}
In this work, we introduced Loss-Guided GFlowNets (LGGFN), a simple yet effective auxiliary training strategy for GFlowNets. Through extensive experiments across four distinct domains, we demonstrated that LGGFN significantly improves exploration efficiency and sample diversity, with improvements ranging from 10\% in complex tasks to over 99\% in sparse reward settings.

While LGGFN shows strong empirical results, computing the auxiliary reward requires evaluating the main model's loss, which adds computational overhead during training. Additionally, the method's performance depends on the choice of loss function and the balance coefficient $\lambda$, though we found it to be relatively robust to these choices in practice.

Future work could explore using alternative signals such as $P_T(x)Z - R(x)$ to guide exploration, though computing marginal distributions remains computationally expensive. Architectural innovations, such as sharing the backbone between main and auxiliary agents while varying only the final layer, may also reduce computational costs while maintaining performance benefits.

\bibliography{aaai2026}

\newpage
\null
\newpage

\section{Appendix / Supplemental Material}

\subsection{A soft introduction to GFlowNets}
The objective of Generative Flow Networks (GFlowNets) is to sample from a state space $\mathcal{X}$ proportionally to a given non-negative reward function $R$. Formally, the desired distribution is defined as:
\[
p(x) = \frac{R(x)}{\sum_{y \in \mathcal{X}} R(y)}.
\]
However, this distribution becomes intractable for large state spaces due to the computational complexity introduced by the summation over all elements of $\mathcal{X}$.

To address this challenge, we consider the scenario where $\mathcal{X}$ is embedded within a larger space $\mathcal{S}$, whose elements can be constructed compositionally (e.g., sentences constructed sequentially by words, molecules built incrementally by atoms). Under this setting, GFlowNets aim to learn transition probabilities between states such that the induced marginal distribution over $\mathcal{X}$ aligns with the reward-proportional distribution $p(x)$.

\begin{figure}[htbp]
  \centering
  \begin{tikzpicture}[
    state/.style={
        circle,
        draw=green!60,
        fill=green!20,
        inner sep=8pt,
        minimum size=1.2cm,
        font=\large\bfseries,
        drop shadow={shadow xshift=2pt, shadow yshift=-2pt, opacity=0.3}
    },
    start/.style={
        state,
        fill=gray!20,
        draw=gray!60,
        line width=2pt
    },
    final/.style={
        state,
        fill=green!20,
        draw=green!60,
        double,
        double distance=2pt
    },
    edge/.style={
        -Stealth,
        line width=1.5pt,
        color=black!70
    },
    edge label/.style={
        font=\normalsize\bfseries,
        fill=white,
        inner sep=3pt,
        rounded corners=2pt,
        draw=black!30,
        opacity=0.9
    },
    value badge/.style={
        circle,
        draw=orange!60,
        fill=orange!20,
        inner sep=2pt,
        minimum size=0.6cm,
        font=\small\bfseries,
        drop shadow={shadow xshift=1pt, shadow yshift=-1pt, opacity=0.2}
    }
]

\node[start] (start) at (0,0) {$-$};
\node[start] (s0) at (3,1.5) {$0$};
\node[start] (s1) at (3,-1.5) {$1$};
\node[final] (s00) at (7,3) {$00$};
\node[final] (s01) at (7,1) {$01$};
\node[final] (s10) at (7,-1) {$10$};
\node[final] (s11) at (7,-3) {$11$};


\node[value badge] at ([yshift=0.8cm] s00) {$3$};
\node[value badge] at ([yshift=0.8cm] s01) {$1$};
\node[value badge] at ([yshift=0.8cm] s10) {$5$};
\node[value badge] at ([yshift=0.8cm] s11) {$2$};

\draw[edge] (start) to[bend left=10] (s0);
\draw[edge] (start) to[bend right=10] (s1);

\draw[edge] (s0) to[bend left=15] (s00);
\draw[edge] (s0) to (s01);

\draw[edge] (s1) to[bend left=15] (s10);
\draw[edge] (s1) to[bend right=15] (s11);

\end{tikzpicture}
  \caption{Example illustrating a simple environment structured as a tree. Nodes represent states, and edges represent actions. Green nodes are terminal and yellow badges represent the reward.}
  \label{fig:tree_example}
\end{figure}

When the environment forms a tree structure with terminal states $\mathcal{X}$ represented by the leaves, determining appropriate transition probabilities is straightforward. Starting from leaf nodes, one propagates values backward by assigning each internal node a value equal to the sum of its children's values, where the leaf node values correspond directly to their rewards. Normalizing each child's value by the parent node's value yields transition probabilities that induce the desired marginal distribution proportional to $R$. For instance, as depicted in Figure~\ref{fig:tree_example}, the value at state 0 would be 4, and the value at the initial (empty) state would be 11.

However, this straightforward approach is limited to tree structures. In general scenarios, the environment forms a Directed Acyclic Graph (DAG), complicating probability assignments.

\begin{figure}[htbp]
  \centering
  \begin{tikzpicture}[
    state/.style={
        circle,
        draw=green!60,
        fill=green!20,
        inner sep=8pt,
        minimum size=1.2cm,
        font=\large\bfseries,
        drop shadow={shadow xshift=2pt, shadow yshift=-2pt, opacity=0.3}
    },
    start/.style={
        state,
        fill=green!20,
        draw=green!60,
        line width=2pt
    },
    final/.style={
        state,
        fill=green!20,
        draw=green!60,
        double,
        double distance=2pt
    },
    edge/.style={
        -Stealth,
        line width=1.5pt,
        color=black!70
    },
    edge label/.style={
        font=\normalsize\bfseries,
        fill=white,
        inner sep=3pt,
        rounded corners=2pt,
        draw=black!30,
        opacity=0.9
    },
    value badge/.style={
        circle,
        draw=orange!60,
        fill=orange!20,
        inner sep=2pt,
        minimum size=0.6cm,
        font=\small\bfseries,
        drop shadow={shadow xshift=1pt, shadow yshift=-1pt, opacity=0.2}
    }
]

\node[state] (start) at (0,0) {$-$};
\node[state] (s0) at (3,1.5) {$0$};
\node[state] (s1) at (3,-1.5) {$1$};
\node[final] (s00) at (7,3) {$00$};
\node[final] (s01) at (7,1) {$01$};
\node[final] (s10) at (7,-1) {$10$};
\node[final] (s11) at (7,-3) {$11$};

\node[value badge] at ([yshift=0.8cm] start) {$4$};
\node[value badge] at ([yshift=0.8cm] s0) {$3$};
\node[value badge] at ([yshift=0.8cm] s1) {$1$};
\node[value badge] at ([yshift=0.8cm] s00) {$3$};
\node[value badge] at ([yshift=0.8cm] s01) {$1$};
\node[value badge] at ([yshift=0.8cm] s10) {$5$};
\node[value badge] at ([yshift=0.8cm] s11) {$2$};

\draw[edge] (start) to[bend left=10] (s0);
\draw[edge] (start) to[bend right=10] (s1);

\draw[edge] (s0) to[bend left=15] (s00);
\draw[edge] (s0) to (s01);

\draw[edge] (s1) to[bend left=15] (s10);
\draw[edge] (s1) to[bend right=15] (s11);

\draw[edge] (s0) to[bend left=20] (s10);
\draw[edge] (s1) to[bend right=20] (s01);

\end{tikzpicture}
  \caption{Example of a DAG environment. Multiple trajectories can reach identical terminal states, and internal states themselves may serve as terminal states if assigned a reward.}
  \label{fig:dag_example}
\end{figure}

In scenarios such as the DAG shown in Figure~\ref{fig:dag_example}, multiple trajectories may converge to a single terminal state, and some internal states can themselves be terminal states with associated rewards. To extend the tree-based approach to DAG environments, the concept of \emph{Flow} is employed. A Flow is a function of the edges adhering to the \textit{flow matching condition}:
\[
F_{\text{out}}(s) + R(s) = F_{\text{in}}(s),
\]
where $F_{\text{out}}(s)$ and $F_{\text{in}}(s)$ are the sum of outgoing and ingoing flows from the node $s$, respectively, and $R(s)$ is $0$ for internal nodes (when $s \notin \mathcal{X}$).
This flow-based representation generalizes the value assignments used in tree structures, and it can be proven that the resulting marginal distribution induced over the state space $\mathcal{X}$ remains proportional to the reward function $R$~\cite{bengio2021flow}.\\

Furthermore, the backward propagation from leaf nodes can alternatively be formulated as a dynamic programming procedure~\cite{lahlou2023advances}. Nevertheless, this approach quickly becomes computationally inefficient when applied to large state spaces. To address this limitation, GFlowNets leverage the generalization capabilities of deep neural networks by parameterizing the flow function as $F_{\theta}$, where $\theta$ represents learnable parameters. Training then involves optimizing these parameters to minimize a loss function that implies the flow matching condition.

\subsection{Theoretical Insights and Curriculum Perspective}
\label{apx:theory}

\paragraph{Stationary points and gradient variance.}
For the trajectory-balance objective 
$\mathcal{L}_{TB}(\tau;\theta)
   =(\log Z_\theta+\sum_t\log P_F^\theta
     -\log R(x_\tau)-\sum_t\log P_B^\theta)^2$,
any parameter configuration that achieves 
$\mathcal{L}_{TB}(\tau)=0$ on the \emph{visited} trajectories
is a stationary point, even if unvisited modes remain.
This explains on-policy mode collapse:
gradients vanish on the sampled support only.
By contrast, sampling from
$q_{\text{aux}}(\tau)\propto R(x_\tau)
   +\lambda\,\mathcal{L}_{TB}(\tau)$
guarantees non-zero mass on high-loss trajectories,
removing such partial-support fixed points.
From an importance-sampling viewpoint, 
LGGFN approximates the variance-minimizing proposal
$q^*(\tau)\propto\|g(\tau)\|$,
since the gradient norm $\|g(\tau)\|$ increases with $\mathcal{L}_{TB}$.
This reduces gradient variance and accelerates convergence.

\paragraph{Implicit curriculum learning.}
Because the loss term naturally decreases during training,
LGGFN also defines a self-paced curriculum:
early in training, the auxiliary GFlowNet focuses on
high-loss (hard) trajectories, encouraging broad exploration;
as learning progresses and losses shrink,
sampling gradually shifts toward the true reward distribution.
This mirrors ideas from curriculum learning 
\citep{bengio2009curriculum, matiisen2019teacher, willems2020mastering},
where tasks are presented in an adaptive order of difficulty.
Here, the “difficulty” of a trajectory is quantified by
the current model’s loss, so the curriculum emerges
automatically without manual task design.

\paragraph{Relation to prior work.}
While \citet{kim2025adaptive} employ a multiplicative loss–reward coupling,
our additive formulation yields the same fixed points,
is numerically more stable,
and preserves a clearer interpretation as both 
importance-weighted sampling and implicit curriculum scheduling.

\subsection{SAGFN and adaptive teachers}
\begin{itemize}
    \item SAGFN \cite{madantowards}: A second agent trains with an intrinsic reward that comes from Random Network Distillation (RND). RND provides exploration bonuses by measuring the prediction error of a randomly initialized target network, encouraging the agent to visit novel states.
    \item Adaptive Teachers \cite{kim2025adaptive}: The second agent trains with the following reward:
\begin{align*}
\log R_{\text{aux}}(x) 
&= \log R^{\text{weighted}}_{\text{aux}}(x; \theta) 
  + \alpha \log R(x) \\
\log R^{\text{weighted}}_{\text{aux}}(x; \theta) 
&= \log\Big(
  \epsilon 
  + \big(1 + C \cdot \mathbb{I}[\delta(\tau;\theta) > 0] \big) \\
&\quad \cdot \delta(\tau;\theta)^2 \Big) \\
\delta(\tau;\theta) 
&= \log R(x) + \log P_B(\tau \mid x) \\
&\quad - \log Z_\theta - \log P_F(\tau; \theta)
\end{align*}

\end{itemize}
\subsection{Related Work}
\label{sec:Related_work}

\textbf{Exploration in Reinforcement Learning:} Exploration is a critical aspect of reinforcement learning (RL). Traditional methods, such as $\varepsilon$-greedy \citep{sutton1998reinforcement}, optimism-based upper confidence bounds (UCB) \citep{auer2002finite}, and Thompson sampling \citep{thompson1933likelihood}, often struggle in complex environments. Approaches based on counts and pseudo-counts \citep{bellemare2016unifying} reduce the focus on frequently visited states, but they fail to scale in high-dimensional settings. Other intrinsic motivation (IM) methods leverage indirect novelty signals as intrinsic rewards. For instance, Random Network Distillation \citep{burda2018exploration} uses the discrepancy between a target and a trainable neural network to capture novelty. instead of absolute novelty, NovelD \citep{zhang2021noveld} uses novelty difference to give large intrinsic reward at the boundary between explored and unexplored regions. Go-Explore \citep{ecoffet2019go} takes a different approach by storing promising states to facilitate easy returns to these regions. Other approaches also used an auxiliary agent to help the main learner improve  \citep{NEURIPS2020_172ef5a9, heinrich2016deep, sukhbaatar2018intrinsic}.

\textbf{Training GFlowNets:} GFlowNets were introduced by \citep{bengio2021flow}, employing the flow-matching objective to reformulate the flow-matching and reward-matching constraints as a loss function. To improve credit assignment and stabilize training, various alternative objectives have been proposed. The detailed-balance objective \citep{JMLR:v24:22-0364} operates at the level of individual transitions, while the trajectory-balance objective \citep{malkin2022trajectory} evaluates entire trajectories. Additionally, the SubTB($\lambda$)-objective \citep{madan2023learning} enables learning from incomplete trajectories, boosting training efficiency. Recent work suggest that weighting the DB objective based on the number of terminal descendants improves the training process \citep{silva2025when}.

\textbf{Exploration in GFlowNets:} The success of GFlowNet training largely depends on the quality of the sampled trajectories. If training is focused on trajectories that consistently converge to a restricted region, sampling diversity can suffer. To address this, \citep{pan2023generative} introduced intrinsic rewards based on novelty methods \citep{burda2018exploration, pathak2017curiosity} to foster exploration of diverse states. Expanding on this idea, \citep{madantowards} proposed a secondary GFlowNet, driven by intrinsic rewards, to help the primary GFlowNet better approximate the target distribution. DGFN \citep{lau2023dgfn} is another approach that involves an auxiliary GFlowNet that is responsible for sampling trajectories, similar to DDQN \citep{van2016deep}. In an effort to enhance exploration, \citep{rector2023thompson} adapted Thompson sampling for GFlowNets. Another approach, suggested by \citep{kim2023local}, involves refining training trajectories using a local search algorithm before presenting them to the GFlowNet for learning.

\subsection{Experiments}
All experiments are run on 3 different seeds and were done using \texttt{torchgfn}~\cite{lahlou2023torchgfn}. For all learnable parameters, the learning rate is set to 0.001, except for the learnable scalar $\log Z$ used in trajectory balance objective, whose learning rate is set to 0.1. After many experiments, this choice seems to work best in general, allowing SAGFN to achieve better results than the ones presented in the original paper \citep{madantowards}. For each GFlowNet, we use the same backbone neural network to learn the forward and backward policy, only changing the final layer to differentiate between them. The auxiliary GFlowNet uses a different backbone network than the main GFlowNet.

\subsubsection{Hypergrid:}
\label{apx:hypergrid}
The Hypergrid environment is traditionally trained with the following reward function:
\begin{align*}
R(x) =\ & R_0 + R_1 \prod_i \mathbb{I}\left(0.25 < \left|\frac{x_i}{H} - 0.5\right|\right) \\
       & + R_2 \prod_i \mathbb{I}\left(0.3 < \left|\frac{x_i}{H} - 0.5\right| < 0.4\right)
\end{align*}

\begin{figure}[H]
  \centering
  \includegraphics[width=0.28\textwidth]{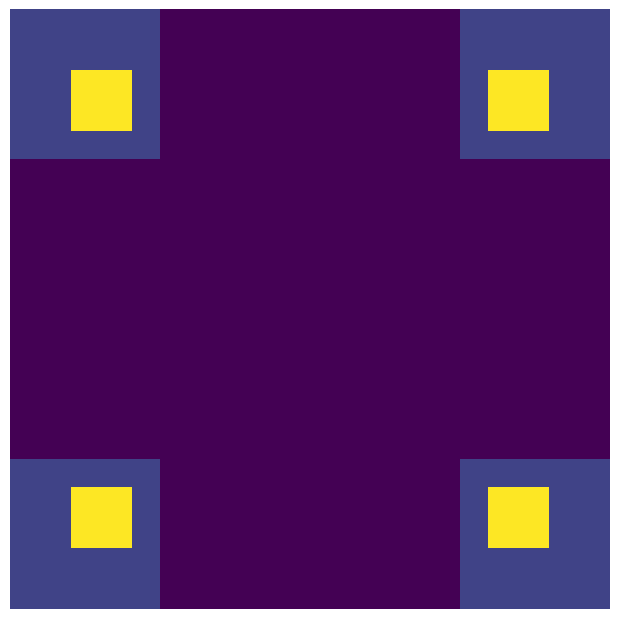}
  \caption{Legend: A 2D Hypergrid environment. States correspond to points $(x_i)_{1 \leq i \leq n}$ on the grid, and actions represent forward transitions in one direction.}
  \label{fig:hypergrid}
\end{figure}

We evaluate the task using grid sizes $H \in \{32, 64, 80, 96, 128, 144\}$, across both 2D and 4D configurations, and vary the reward sparsity using $R_0 \in \{0.01, 0.001, 0.0001, 0.00001\}$. The agent is parameterized by a multi-layer perceptron (MLP) consisting of two hidden layers with 256 units each and ReLU activations. After extensive tuning of the regularization hyperparameter $\lambda$, we observed that performance remains consistent as long as $\lambda \neq 0$; all results reported in the paper use $\lambda = 1$.

For SAGFN, we use the hyperparameters from original work \cite{madantowards}:
\[
\beta_{e}^{\text{main}} = 1,\quad \beta_{e}^{\text{aux}} = 0.25,\quad \beta^{\text{aux}} = 1,\quad \beta^{\text{main}} = 1,\quad \beta_i = 1
\]

To compare LGGFN against the Adaptive Teachers from \cite{kim2025adaptive}, we use the same hyperparameters as in~\cite{kim2025adaptive}, where $C = 19$ and $\alpha = 0.5$.

\vspace{-0.2cm}

\subsubsection{Visualization of HyperGrid learned distributions:}

\begin{figure}[h]
    \centering
    \includegraphics[width=0.46\textwidth]{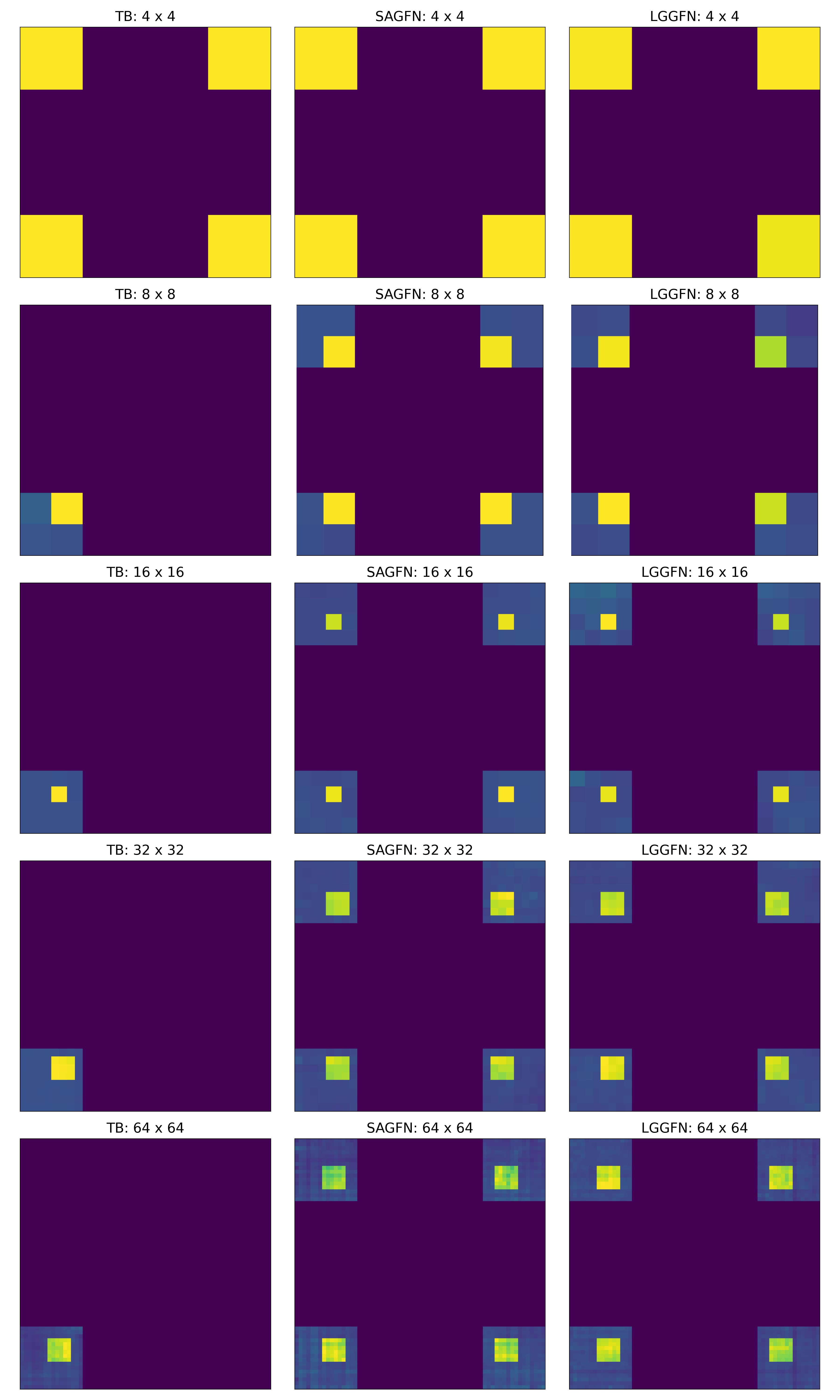}
    \caption{Learnt distributions on environment with size from 4 to 64.}
    \label{fig:hypergrid_examples_1}
\end{figure}
\begin{figure}[!t]
    \centering
    \includegraphics[width=0.46\textwidth]{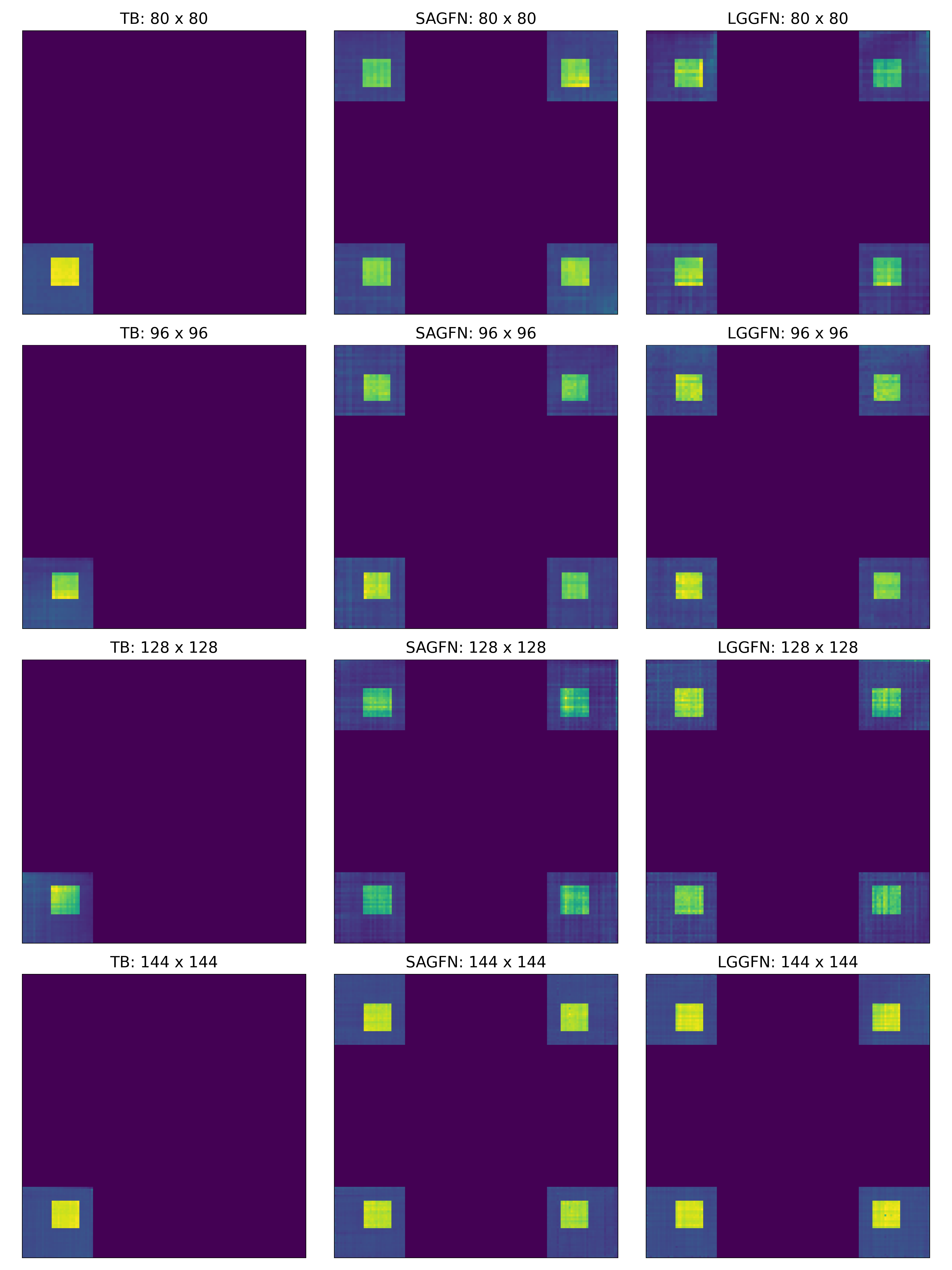}
    \caption{Learned distributions on environment with size from 80 to 144.}
    \label{fig:hypergrid_examples_2}
\end{figure}

\subsubsection{Valid sequences:}
\label{apx: sequences}
\begin{table}[H]

\label{tab:catalan-numbers}
\centering
\begin{tabular}{@{}cc@{}}
\toprule
\textbf{N} & \textbf{Catalan Number} \\
\midrule
8  & 1430 \\
12 & 2674440 \\
16 & 35357670 \\
20 & 6564120420 \\
24 & 1289904147324 \\
\bottomrule
\end{tabular}
\caption{Catalan numbers for selected values of \( N \).}

\end{table}

This task involves generating a binary sequence by autoregressively appending individual bits. Compared to the original \textsc{BitSequences} environment, several key modifications make this version more challenging:

\begin{itemize}
    \item \textbf{Reward Function}: In contrast to the original setting, which featured unstructured and near-random rewards, the modified environment introduces a structured and sparse reward function. Substantial rewards are allocated exclusively to sequences that are both complete and balanced, while the vast majority of other sequences yield negligible reward. Exploration can be made even more challenging by assigning to small sequences (e.g., those shorter than 4 or 8 bits) deceptively low but nonzero rewards, i.e., one that is higher than those of arbitrary incomplete sequences but significantly lower than optimal ones. This design introduces local optima (without affecting much the marginal distribution of the problem) that can mislead inadequately exploratory training strategies, potentially trapping the policy and mimicking the behavior illustrated in Figure~\ref{fig:chain_plot}.
    
    \item \textbf{Terminating States}: In the original environment, only fully constructed sequences are terminal. Here, the agent may choose an exit action at any sequence length. This change increases task complexity, as the agent must learn to discern whether a partial sequence is promising enough to continue or whether to exit, despite most high-reward sequences occurring only at full length.
    
    \item \textbf{Word Size}: In the original experiments, sequences had fixed length, and each step appended a binary word of size \( k \), varying across experiments. In our version, sequences terminate upon taking the exit action, and the agent adds one bit at a time. This increases the relative importance of the exit action, which now has a probability of \( 1/3 \) (compared to \( 1/(2^k + 1) \) when adding \( k \)-bit words), making the decision to stop a more frequent and significant consideration.
\end{itemize}

For this task, we use \texttt{GPT2Model} from Hugging Face's transformers library \cite{wolf2019huggingface} with 3 hidden layers, 8 attention heads, and embeddings of dimension 64. For LGGFN, $\lambda$ was set to $1$ and for SAGFN, $\{
\beta_{e}^{\text{main}} = 1, \beta_{e}^{\text{aux}} = 1, \beta^{\text{aux}} = 1, \beta^{\text{main}} = 1, \beta_i = 1
\}$.

\begin{table*}[t]
\centering

\setlength{\tabcolsep}{4pt} 
\renewcommand{\arraystretch}{1.2} 
\begin{tabular}{@{}l>{\columncolor{lightgray}}c c>{\columncolor{lightgray}}c c>{\columncolor{lightgray}}c c>{\columncolor{lightgray}}c c>{\columncolor{lightgray}}c}
\toprule
\textbf{\# Nodes} & 5 & 6 & 7 & 8 & 9 & 10 & 11 & 12 & 13 \\
\midrule
\textbf{\# DAGs} 
& $2.93{\times}10^4$
& $3.78{\times}10^6$
& $1.14{\times}10^9$
& $7.84{\times}10^{11}$
& $1.21{\times}10^{15}$
& $4.18{\times}10^{18}$
& $3.16{\times}10^{22}$
& $5.22{\times}10^{26}$
& $1.87{\times}10^{31}$ \\
\bottomrule
\label{tab:number-of-dags}
\end{tabular}
\caption{Approximate number of Directed Acyclic Graphs (DAGs) for different node counts}

\end{table*}

\subsubsection{Bayesian Structure Learning:}
\label{apx:graphs}
\paragraph{Markov Equivalence:}
Two graphs $G_1$ and $G_2$ are said to be Markov equivalent if they encode the same set of conditional independencies (Figure \ref{fig:markov_eq}):

\begin{gather*}
G_1 \sim G_2 \;\Leftrightarrow\; CI(G_1) = CI(G_2) \\
\text{where} \quad CI(G) = \{ (X, Y, Z) \mid X \perp\!\!\!\perp Y \mid Z \text{ in } G \}
\end{gather*}

\begin{figure}[H]
    \centering
    \resizebox{0.2\textwidth}{!}{\begin{tikzpicture}[->, >=latex, line width=2.5pt, node distance=2.5cm and 2.5cm]

  \tikzstyle{Xnode} = [circle, draw, fill=red!20, minimum size=50pt, inner sep=0.1pt]
  \tikzstyle{Ynode} = [circle, draw, fill=green!20, minimum size=50pt, inner sep=0.1pt]
  \tikzstyle{Znode} = [circle, draw, fill=blue!20, minimum size=50pt, inner sep=0.1pt]

  \node[Xnode] (X1) at (0, 0) {\LARGE\textbf{X}};
  \node[Znode] (Z1) [right=of X1] {\LARGE\textbf{Z}};
  \node[Ynode] (Y1) [right=of Z1] {\LARGE\textbf{Y}};
  \draw (X1) -- (Z1);
  \draw (Z1) -- (Y1);
  \node[left=0.6cm of X1] {\Huge \(G_1\)};

  \node[Xnode] (X2) at (0, -2.5) {\LARGE\textbf{X}};
  \node[Znode] (Z2) [right=of X2] {\LARGE\textbf{Z}};
  \node[Ynode] (Y2) [right=of Z2] {\LARGE\textbf{Y}};
  \draw (Z2) -- (X2);
  \draw (Z2) -- (Y2);
  \node[left=0.6cm of X2] {\Huge \(G_2\)};

  \node[Xnode] (X3) at (0, -5) {\LARGE\textbf{X}};
  \node[Znode] (Z3) [right=of X3] {\LARGE\textbf{Z}};
  \node[Ynode] (Y3) [right=of Z3] {\LARGE\textbf{Y}};
  \draw (Y3) -- (Z3);
  \draw (Z3) -- (X3);
  \node[left=0.6cm of X3] {\Huge \(G_3\)};

\end{tikzpicture}}
    \caption{Three Markov equivalent DAGs.}
    \label{fig:markov_eq}
\end{figure}

For this environment, we first generate a random directed acyclic graph (DAG) (by sampling an Erdős–Rényi graph and assigning edge directions such that the source node has a lower index than the target node). Data is then generated according to a linear-Gaussian Structural Causal Model (SCM):
\[
X_i = \sum_{j \in \text{Pa}(i)} X_j + \varepsilon_i, \quad \varepsilon_i \sim \mathcal{N}(0, \sigma^2),
\]
where $\text{Pa}(i)$ denotes the set of parents of node $i$ in the DAG, and summation over an empty set is defined to be zero.

In this setting, the states of the environment correspond to DAGs, while the actions consist of adding valid edges, i.e., those that preserve acyclicity. The reward function for a given graph is defined based on the graph itself and its parent graphs in the state space:
\[
\log R(G) = \sum_{j=1}^{d} \text{LocalScore}(X_j \mid \text{Pa}_G(X_j)),
\]
where $\text{LocalScore}$ denotes the BGe score \citep{heckerman1995learning}, and $\text{Pa}_G(X_j)$ represents the set of parents of node $X_j$ in graph $G$. For this task, we use a directional graph-based neural architecture to compute action logits for graph generation. The model supports decisions for both edge creation between nodes and an optional stop action that signals the end of generation. The network begins by embedding node identifiers into a continuous feature space. These embeddings are processed through multiple layers of direction-aware graph convolutions that distinguish between incoming and outgoing edges. Each layer includes residual connections to preserve information across depths and applies layer normalization to stabilize training. The output of each convolutional layer consists of separate representations for incoming and outgoing directional features. These are refined independently using small feedforward networks and then recombined into a unified node representation. To produce edge-level action scores, the model computes dot products between the feature vectors of source and target nodes. In addition, a pooled summary of all node embeddings can be used to compute a score for the optional stop action. The final output consists of unnormalized logits representing the desirability of all possible edge additions and, if applicable, the decision to terminate the graph construction process.

Unlike the previous two tasks, this task exhibits significantly higher loss values, often exceeding the magnitude of the actual rewards. Moreover, the loss values vary with the number of nodes in the input graph. To account for this, the hyperparameter $\lambda$ was selected individually for each graph size by empirically observing the loss range over a series of iterations and then choosing a value of $\lambda$ that brings the loss scale closer to that of the reward.
For the SAGFN model, we adopt the following configuration of weighting coefficients $\{
\beta_{e}^{\text{main}} = 1,\quad \beta_{e}^{\text{aux}} = 1,\quad \beta^{\text{aux}} = 1,\quad \beta^{\text{main}} = 1,\quad \beta_i = 1
\}$. This setup ensures that the intrinsic reward remains within the same range as the extrinsic reward, facilitating more stable training.

\subsubsection{mRNA Sequence Design:}
\label{sec:codondesign}
Messenger RNA (mRNA) sequences are composed of codons, triplets of nucleotides, each of which encodes a specific amino acid. Due to the redundancy of the genetic code, most amino acids can be encoded by multiple synonymous codons. Thus, a single protein sequence can correspond to many possible mRNA sequences. Leveraging this redundancy, we introduce the \textsc{CodonDesign} environment to model codon-level mRNA sequence design conditioned on a fixed protein sequence. States represent partially constructed mRNA sequences, and actions correspond to choosing synonymous codons or a special exit action. At each position $t$, only codons that encode the $t$-th amino acid are permitted. The episode terminates after $T$ steps (the protein length), followed by the exit action. The reward function is multi-objective and reflects the quality of the generated mRNA according to three criteria:
\begin{enumerate}
    \item \textbf{GC-content:} The fraction of G or C nucleotides in the sequence. High GC-content generally correlates with increased mRNA stability and translation efficiency \cite{courel2019gc}, in particular, optimizing GC-content has a similar effect to optimizing codon usage \cite{zhang2023algorithm}.
    \item \textbf{Minimum free energy (MFE):} The (negative) folding free energy of the mRNA's secondary structure. A lower MFE (more negative) indicates a more stable folded structure. 
    \item \textbf{Codon adaptation index (CAI):} A measure of how well the sequence conforms to species-specific codon usage preferences. Higher CAI typically leads to more efficient translation \cite{sharp1987codon}.
\end{enumerate}
These components are linearly combined:
\[
R(s) = w_1 \cdot \text{GC}(s) - w_2 \cdot \text{MFE}(s) + w_3 \cdot \text{CAI}(s),
\]
where $w \in \mathbb{R}^3$ are user-defined weights. This reflects the intuition from prior work that mRNA design must balance structural stability and codon optimality. However, these objectives are often in tension, making diversity and mode coverage essential for identifying high-quality trade-offs in the design space.
\noindent Across the three GFlowNet variants, TB, SAGFN, and LGGFN, we generated diverse mRNA sequences encoding the same target protein. Despite the shared amino acid sequence, the models produced variants that varied significantly in their codon usage, GC content, and predicted RNA stability by the MFE metric. Notably, LGGFN consistently yielded sequences with high Codon Adaptation Index (CAI) and GC content, reflecting improved translation potential and structural stability. Furthermore, analysis of pairwise Levenshtein distances between generated sequences revealed substantial diversity, especially in LGGFN and SAGFN outputs. This diversity suggests that the models are exploring multiple high-reward regions of the vast mRNA sequence space.

\end{document}